\pdfoutput=1

\documentclass[11pt]{article}

\usepackage{acl}

\usepackage{times}
\usepackage{latexsym}
\usepackage{booktabs}
\usepackage[T1]{fontenc}

\usepackage[utf8]{inputenc}

\usepackage{microtype}
\usepackage{multirow}
\usepackage{graphicx}
\usepackage{amsmath}
\usepackage{amssymb}
\newcommand{\ourmethod}{\textsc{SMDT}}
\title{SMDT: Selective Memory-Augmented Neural Document Translation}

\author{
 Xu Zhang\textsuperscript{\rm 1 \thanks{\ Contribution during internship at Microsoft Research Asia.}}, 
 Jian Yang\textsuperscript{\rm 2}, 
 Haoyang Huang\textsuperscript{\rm 3}, 
 Shuming Ma\textsuperscript{\rm 3}, 
 Dongdong Zhang\textsuperscript{\rm 3},
\\ \textbf{Jinlong Li}\textsuperscript{\rm 1},
 \textbf{Furu Wei} \textsuperscript{\rm 3},
 \\ 
 \textsuperscript{\rm 1}University of Science and Technology of China \\
 \textsuperscript{\rm 2}Beihang University \\
 \textsuperscript{\rm 3}Microsoft Research Asia \\
 xuz@mail.ustc.edu.cn; jiaya@buaa.edu.cn; \\ \{shumma, haohua, dozhang, fuwei\}@microsoft.com; jlli@ustc.edu.cn}

\begin{document}
\maketitle
\begin{abstract}

Existing document-level neural machine translation (NMT) models have sufficiently explored different context settings to provide guidance for target generation.
However, little attention is paid to inaugurate more diverse context for abundant context information. In this paper, we propose a \textbf{S}elective \textbf{M}emory-augmented Neural \textbf{D}ocument \textbf{T}ranslation model (\ourmethod{}) to deal with documents containing large hypothesis space of the context. Specifically, we retrieve similar bilingual sentence pairs from the training corpus to augment global context and then extend the two-stream attention model with selective mechanism to  capture local context and diverse global contexts. This unified approach allows our model to be trained elegantly on three publicly document-level machine translation datasets and significantly outperforms previous document-level NMT models.


\end{abstract}

\section{Introduction}
Neural Machine Translation (NMT) has achieved great progress in sentence-level translation \cite{Bahdanau2015NeuralMT,xia2017deliberation,vaswani2017attention}. However, sentence-level models may ignore discourse phenomena and translate sentences in isolation.
The use of context in document-level translation has been advocated by NMT pioneers for decades and can be divided into two categories.
The first uses several surrounding sentences as the context ~\cite{miculicich2018document,voita2019good,ma2020simple,xu2020efficient},
the other uses the whole document context
~\cite{maruf2018document,tan2019hierarchical,xiong2019modeling,zheng2020towards}. 
But these contexts are limited to a fixed span of the document and how to increase the diversity of context is still a challenge.

Inspired by memory-based methods in sentence-level translation \cite{gu2018search,xu2020boosting,shang2021guiding}, we can explicitly retrieve similar sentences to augment document-context. But \citet{bao2021g} conducted a detailed analysis that translating longer document may fail to not converge or stick around local minima. Besides, more document-contexts may contain more irrelevant words. Therefore, how to effectively leverage knowledge and select the proper words from a larger hypothesis space is another challenge.


To address the aforementioned problems, we propose a \textbf{S}elective \textbf{M}emory-augmented Neural \textbf{D}ocument \textbf{T}ranslation model (\ourmethod{}). The \ourmethod{} firstly retrieves sentence pairs from training corpus to enrich translation memory (TM). We merge the entire document and the retrieval content into one input unit and divide it by sentence. To capture the short- and long-term dependencies, we introduce two-stream attention with selective mechanism: 1) For each sentence, local attention can empower the model to capture the short-term dependencies and block other sentences, 2) and diverse global attention adopts three types of attention to obtain different aspects of context including retrieved sentence pairs, all sentences of the entire document and the most adjacent sentences.
Following \cite{yang2021learning}, we add a selection mechanism at the top of the encoder block to reduce 
redundant words from TM. 
In addition, sentence-level translation task has been shown can also help document-level translation task \cite{bao2021g}. To verify the idea, we perform multi-task learning and achieve significant improvement.

Our contributions can be summarized as follows: 
(i) We introduce \ourmethod{}, a novel method of augmenting translation memory with retrieved similar sentences and including two-stream attention with selective mechanism to capture appropriate knowledge from multiple contextual information.
(ii) We evaluate this model on three document-level machine translation datasets. The experimental results show that our model can achieve competitive performance compared to previous works. We also present ablation studies to provide useful insights on the effectiveness of model variants.

\section{Method}
\subsection{Problem Definition}
Given a source document $X = \left\{x_1,\dots,x_m\right\}$ and target document $Y=\left\{y_1,\dots,y_m\right\}$, we can retrieve a set of similar sentence pairs $Z_X=\left\{z_1,\dots,z_m\right\}$ using similarity metric, 
where each pair is combined with the retrieved source and target sentence. The target translation probability is formulated as below:
\begin{SmallEquation}
\begin{equation} \label{translaiton-probability}
    P(Y|X) = \prod \limits_{i=1}^m P(y_i|X,Z_X;\theta)
\end{equation} 
\end{SmallEquation}where $\theta$ are model parameters.

\subsection{Input Construction}
We first use the widely-used search engine Lucene\footnote{\url{https://github.com/apache/lucene-solr}} to retrieve similar sentences, which provide diverse context candidates. Given a source sentence $x$, the similar source sentence $x_{i}^{r}$ and corresponding target sentence $y_{i}^{r}$ from the bilingual training corpus are concatenated as one part $z_{i}=(x_{i}^{r},y_{i}^{r})$. The input of our work is set as the concatenated source sentences and the corresponding translation pairs $[X;Z]$, where we use a special separator to distinguish the original document and the retrieval sentences.

\subsection{Two-Stream Attention with Selective TM}
Self-attention and cross-attention with multi-head are used to obtain information from different representation subspaces at different positions \cite{vaswani2017attention}. Each head corresponds to a scaled dot-product attention, which operates on the query $Q$, key $K$, and value $V$. The masking matrix $M$ is used to mask illegal tokens before softmax:
\begin{SmallEquation}
\begin{equation} \label{attention}
    \text{Attention}(Q,K,V,M) = \text{softmax}(\frac{QK^T}{\sqrt{d_k}}+M)V
\end{equation} 
\end{SmallEquation}where $d_k$ is the dimension of the key vector.

The output values are concatenated and projected by a feed-forward layer to get final values:
\begin{SmallEquation}
\begin{equation} \label{mha}
\begin{split}
    \text{MHA}(Q,K,V,M) = \text{Concat}(head_1, ..., head_h)W^O \\
     \text{\textit{where} head}_i = \text{Attention}(QW_i^Q, KW_i^K, VW_i^V, M) 
\end{split}
\end{equation} 
\end{SmallEquation}where MHA denotes the multi-head attention with $h$ heads. $W_O$, $W_Q$, $W_K$, and $W_V$ are parameters.

\begin{figure}[t]
\begin{center}
    \includegraphics[width=0.75\columnwidth]{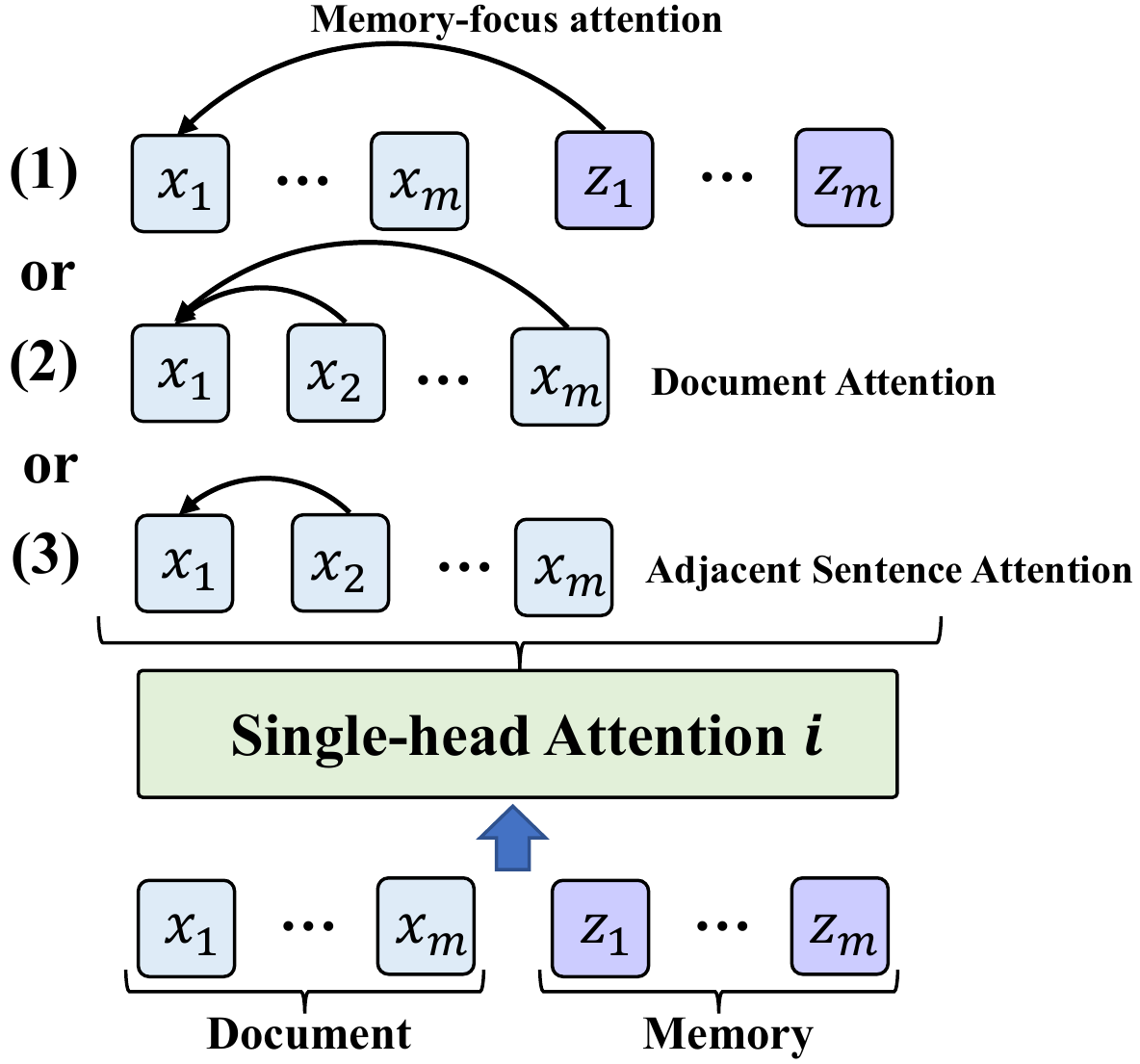}
    \caption{Explanation of the diverse global attention. Each head chooses one attention type from (1)-(3).}
    \label{figure1}
    \vspace{-10pt}
\end{center}
\end{figure}
Our model extends the multi-head attention with two-stream attention architecture including local attention and diverse global attention. 
The local attention is employed to capture local dependencies and diverse global attention at top layers is used for global context information.

\paragraph{Local Attention}
Local attention for encoder and decoder only obtains the local sentence information and disregards other sentences. We modify the masking matrix $M$ in Equation \ref{attention} to discard irrelevant sentences.

\paragraph{Diverse Global Attention}
Since different heads focus on diverse spans of the sentence \cite{sukhbaatar2019adaptive}, we adopt diverse global attention to concentrate on different context ranges. 
More specifically, we provide three types of attention for each head to obtain different aspects of context.
In Figure \ref{figure1}, the diverse global attention includes three parts:
(i) \textbf{Memory-focus attention}, where each source sentence depends on the corresponding retrieved sentence pair. The translation memory can provide additional information and the context of the target locale.
(ii) \textbf{Document Attention}, which simultaneously focuses on all sentences in the same document.  The entire document can provide rich information, but will also introduce a lot of noise.
(iii) \textbf{Adjacent Sentence Attention}, which regards close sentences as the context instead of entire document. Close sentences provide high-relevant context and alleviate the problem of large hypothesis space. In our work, we unify three types of attention as the diverse global attention to extract global and abundant representations from the document and retrieved pairs.
Local attention and diverse global attention are combined using a commonly used gate-sum module \cite{zhang2016gated} at the top two layers.

\paragraph{Selection Mechanism}
Following \citet{yang2021learning}, we add a selection layer at the top of the encoder to select useful pieces of the translation memory.
We use hidden states $h^{L+1}$ of $L$ weighted combinations of the encoder block features to choose the proper words, where the binary classification is used to decide whether the $j$-th position in the retrieved is retained:
\begin{SmallEquation}
\begin{equation} \label{selectlayer}
    \begin{split}
    h^{L+1} &= \text{Self-Attn}(\sum_{i=1}^{L}a_i h_i)\\
    \sigma &= \text{argmax}\ \text{softmax}(W_s h^{L+1})
    \end{split}
\end{equation} 
\end{SmallEquation}where $a_i$ is the learnable weight of the $i$-th block and $W_s\in \mathbb{R}^{d\times 2}$. $\sigma=(\sigma_1,...,\sigma_t)$ with $t$ words. $\sigma_j=1$ denotes the $j$-th word is selected while $\sigma_j=0$ denotes $j$-th word is discarded.

\subsection{Multi-task Learning Framework}
Our multi-task learning framework consists of main and auxiliary tasks. Herein, we refer to the document machine translation task as the main task and the sentence machine translation task as the auxiliary task.
We separately construct the training objectives for sentence-level and document-level translation based on Equation \ref{translaiton-probability}.

\begin{SmallEquation}
\begin{equation} \label{multitask}
    L = \mathbb{E}_{X,Y,Z_{X} \in D_{b}}\log[-P(Y|X,Z_{X};\theta)]
\end{equation} 
\end{SmallEquation}where $\theta$ are model parameters. $D_b$ is the bilingual corpus and $Z_{X}$ are retrieved pairs.

\paragraph{Document Translation}
The main task is translation at the document level. When preprocessing the data, we do not limit the number of sentences in the document, so each instance contains multiple sentences and retrieved translation pairs. 

\paragraph{Sentence Translation}
The auxiliary task is sentence-level translation.
When $m=1$ in Equation \ref{translaiton-probability}, each instance is composed of a single sentence and the corresponding retrieved pair. Compared with the main task, sentence-level translation allows the model to focus on learning the local dependencies under the current single sentence and the external dependencies provided by translation memory, which helps the model learn to select important retrieved pieces.

\begin{table}[t]
\centering
\resizebox{1.0\columnwidth}{!}{
\begin{tabular}{l|ccc}
\toprule
 \textbf{ Method} & {TED} & {News} & {Europarl}\\

\midrule
RNN \cite{bahdanau2015neural} & 19.24 & 16.51 & 26.26\\
SENTNMT \cite{vaswani2017attention} & 23.10  & 22.40  & 29.40 \\
SENTNMT (our implementation) & 24.71  & 25.04  & 31.50   \\
\midrule
HAN \cite{miculicich2018document} & 24.58 & {25.03} & 28.60 \\
SAN \cite{maruf2019selective} & 24.42  & 24.84 & 29.75 \\
Hybrid Context \cite{zheng2020towards} & {25.10}   & 24.91  & 30.40   \\
Flat-Transformer \cite{ma2020simple} & 24.87  & 23.55  & 30.09  \\
G-Transformer \cite{bao2021g} & 23.53 & 23.55 & 32.18 \\ \midrule
\textbf{\ourmethod{} (Our method)} & \textbf{25.12} & \textbf{25.76} & \textbf{32.42} \\
\bottomrule
\end{tabular}}
\caption{Comparison results on three document-level machine translation benchmarks with BLEU metrc.}
\label{main}
\end{table}

\section{Experiments}

\begin{table}[ht]
\centering
    \resizebox{0.75\columnwidth}{!}{
\begin{tabular}{l|lll}
\toprule
Dataset &  \#Sent & \#Documents  \\
\midrule
TED & 0.21M/9K/2.3K & 1.7K/92/22\\
News & 0.24M/2K/3K & 6K/80/154\\
Europarl & 1.67M/3.6K/5.1K & 118K/239/359\\
\bottomrule
\end{tabular}}
\caption{Statistics of three datasets.}
\label{datasets}
\end{table}
\paragraph{Datasets}
Following the previous work \cite{maruf2019selective}, we use three public datasets including TED, News, and Europarl. 
Statistics of these datasets are reported in Table \ref{datasets}. 
The documents are truncated to 1000 tokens.
We use Moses toolkit\footnote{\url{https://github.com/moses-smt/mosesdecoder}} to tokenize the sentences and encode words into subwords \cite{sennrich2016neural} with 30K merge operations. The evaluation metric is the case-sensitive BLEU points \cite{papineni2002bleu}.

\paragraph{Training Setting}

We use 512 embedding size, 2048 FFN size, and 8 attention heads. For diverse global attention, 3 heads use memory-focus attention, 3 heads use document attention, and 2 heads use adjacent sentence attention. Diverse global attention is only used in the top two layers.
A dropout rate of 0.3 is applied to residuals, attentions, and ReLU connections. We use Adam optimizer with $\beta_1=0.9$ and $\beta_2=0.98$ to train all models, and apply label smoothing with an uncertainty of 0.1. All models are trained on 4 GPUs of Nvidia V100. We determine the number of updates/steps automatically by the early stop on
validation set.


\paragraph{Main Results} In Table \ref{main}, we present the BLEU score of our model and other baselines on TED, News, and Europarl.
The results show that our \ourmethod{} can obtain a leading performance of 25.12/25.76/32.42 on three datasets.
It is noticeable that our proposed method significantly outperforms the Transformer-based SENTNMT baseline model on average, which proves our method sufficiently utilizes different types of contextual information.
Compared with the G-Transformer that also uses two-stream attention, our \ourmethod{} has improved significantly on TED and News, partly because of the use of the auxiliary task, and partly because of the improvements in our model architecture.

\section{Analysis}
\paragraph{Ablation Study}
To analyze the effect of each component of our \ourmethod{}, we conduct an ablation study by removing them from our models on the TED dataset in Table \ref{ablation}. 
To eliminate the experimental interference of the auxiliary task, our ablation study is only trained on the main task.
After removing the selection layer, the performance drops by 0.86 BLEU scores, indicating the necessity of introducing the selection mechanism for noise filtering.
After removing the diverse global attention, the performance of the model drops by 1.04 BLEU scores, which shows that only using local attention is not enough for document-level translation.

\begin{table}[ht]
\centering
\resizebox{0.6\columnwidth}{!}{
\begin{tabular}{l|l}
\toprule
Operation & BLEU  \\
\midrule
\textbf{\ourmethod{}} & \textbf{24.55} \\
\midrule
w/o select layer & 23.69 \\
w/o diverse global attention & 23.51\\
\bottomrule
\end{tabular}}
\caption{Ablation study of our model only trained with the document translation task on the TED dataset.}
\label{ablation}
\vspace{-10pt}
\end{table}

\begin{table}[ht]
\centering
\resizebox{0.65\columnwidth}{!}{
\begin{tabular}{l|lll}
\toprule
Context & TED & News & Europarl \\
\midrule
\textbf{diverse (\ourmethod{})} & \textbf{24.55} & \textbf{24.32} & \textbf{32.24} \\
\midrule
only doc & 23.87 & 24.09 & 32.20 \\
only retrieval & 24.17 & 24.28 & 32.02 \\
\bottomrule
\end{tabular}}
\caption{Context replacement results of our model only trained with the document translation task on three datasets, emphasizing importance of document and retrieved sentence pairs.}
\label{replacement}
\vspace{-10pt}
\end{table}

\paragraph{Context Replacement}
To investigate the influence of different contexts on translation quality, we conduct a context replacement experiment for global attention.
We only use the main task for training in the context replacement experiment to avoid the interference of the auxiliary task.
Table \ref{replacement} summarizes the results of the context replacement experiment.
It can be seen that the performance of our \ourmethod{} model on TED and News has declined due to the absence of the auxiliary task, which shows the important role of the sentence translation task.
After replacing the contexts of all heads in global attention with the whole input document, the model drops 0.68/0.23/0.04 on three datasets respectively. 
Similarly, if the contexts of all heads are replaced with retrieved translation memory, the model drops 0.38/0.04/0.22 on three datasets respectively.
The experiment shows that the use of multiple types of contexts, including the context of the target locale and the context of high-relevant content using a smaller hypothesis space, is effective for improving the quality of document translation.

\section{Related Work}
\paragraph{Neural Machine Translation} 
Neural Machine Translation (NMT) has achieved great progress in sentence-level translation \cite{Bahdanau2015NeuralMT,xia2017deliberation,vaswani2017attention}. Context-aware NMT is a more practical task because the sentence to be translated requires supplementary context to address the problem of pronoun translation, lexical cohesion, and discourse connectives \cite{maruf2018document}. 

\paragraph{Document-level Translation}
Previous works extend NMT models with an extra module to encode context \cite{zhang2018improving,maruf2019selective,werlen2018document,yang2019enhancing}. 
They ignore the relationship between current sentences and context since the dual encoder lacks interaction with each other. A unified encoder \cite{ma2020simple} is proposed to jointly encode current sentences and context. Following this line of research, two-stream self-attention \cite{bao2021g,zhang2020long} is adopted to capture long- and short-term dependencies. But these methods using the whole document encounter training difficulties, such as large hypothesis space and low-quality context with much noise. Different from aforementioned methods, our model extends the two-stream self-attention with selective memory and diverse global attention to enrich context information. 

\section{Conclusion}
In this work, we explored solutions to improve the diversity of context in document-level machine translation. We propose a two-stream attention model called \ourmethod{} based on selective memory of retrieved pairs. To utilize useful spans of the document and retrieved pairs, diverse global attention is proposed to provide different aspects of global dependencies, which separately focuses on retrieved sentence pairs, the whole document, and the adjacent sentences. Experimental results demonstrate our model can effectively improve performance on three benchmarks in English$\to$German translation direction.

\bibliographystyle{acl_natbib}
\nocite{*}
\bibliography{anthology}

\end{document}